%% file: main.tex
\documentclass[conferece]{IEEEtran}
\IEEEoverridecommandlockouts
\usepackage{cite}
\usepackage{amsmath,amssymb,amsfonts}
\usepackage{algorithmic}
\usepackage{graphicx}
\usepackage{textcomp}
\usepackage{xcolor}
\usepackage{verbatim}
\usepackage{setspace}
\usepackage{makecell}
\usepackage{graphicx}
\usepackage{subfigure}
\def\BibTeX{{\rm B\kern-.05em{\sc i\kern-.025em b}\kern-.08em
    T\kern-.1667em\lower.7ex\hbox{E}\kern-.125emX}}
\usepackage[justification=centering]{caption}

\begin{document}
\input{0_Abstract.tex} 
\input{1_Introduction}
\input{2_Related_Work}
\input{3_latency}
\input{5_Conclusion}

\bibliographystyle{IEEEtran}
\bibliography{ref}

\end{document}

%% file: 0_Abstract.tex
\title{The Importance of Autonomous Driving Using 5G Technology}

\author{\IEEEauthorblockN{
Yuanzhe Jin,}
\IEEEauthorblockN{Neelanshi Varia,}
\IEEEauthorblockN{Chixiang Wang}
\IEEEauthorblockA{ 
\\Northwestern University, Evanston, IL, USA\\
yzjin@u.northwestern.edu, neelanshiv2@u.northwestern.edu, chixiang@u.northwestern.edu}
}

\maketitle

\begin{abstract}
The three keys to autonomous driving are sensors, data integration, and 100\% safety decisions. In the past, due to the high latency and low reliability of the network, many decisions had to be made locally in the vehicle. This puts high demands on the vehicle itself, which results in the dilatory commercialization of automatic driving. With the advent of 5G, these situations will be greatly improved.

In this paper, we present the improvements that 5G technology brings to autonomous vehicles especially in terms of latency and reliability amongst the multitude of other factors. The paper analyzes the specific areas where 5G can improve for autonomous vehicles and Intelligent Transport Systems in general (ITS) and looks forward to the application of 5G technology in the future.
\end{abstract}


%% file: 1_Introduction.tex
\section{Introduction}

The fifth-generation mobile communication technology (5G) is the latest generation of cellular mobile communication technology, which is an extension of the 4G (LTE) system and fulﬁlls the wireless broadband communication speciﬁcations of 2020 and beyond \cite{agiwal2016next} \cite{gupta2015survey}. 
Over the past three decades, the mobile network has undergone several huge innovations, which has brought a big leap in efficiency and experience. 

In the 1G era, people were liberated from fixed telephones. 2G solved various limitations of 1G. 3G introduced mobile broadband, and 4G brought faster and more stable mobile broadband. In addition to drastically increasing the transmission rate, 5G must also have ultra-high reliability, extremely low latency, ultra-wide coverage, and ultra-high network carrying capacity. 

By achieving these principles, 5G will enable the technologists to perform more complex tasks than ever that were previously limited by computational and latency bottlenecks. The human consumers might not even need the ultra-level speed of 5G but the machines, robots, and vehicles can compute and communicate at a drastically better latency leading to ace in numerous tasks. Based on some consensus, 5G will be mainly applied in three major areas, namely: enhanced mobile broadband (eMBB), mission-critical service (also known as Ultra-Reliable Low Latency Connection (uRLLC)) and Massive IoT.

Compared with the LTE network, due to the application of new technologies such as a self-contained integrated sub-frame and scalable TTI, the delay is significantly reduced in 5G. 5G will support applications and systems that can only tolerate delay lower than 1 millisecond. 

5G is generally considered to be one of the major developmental technologies amongst Blockchain, IoT, Artificial Intelligence, etc. and it will bring huge changes and improvements in the capability of mobile networks, particularly around new radio access technology, higher frequencies usage, network re-architecture.

\begin{figure}[htbp]
  \includegraphics[scale=0.35]{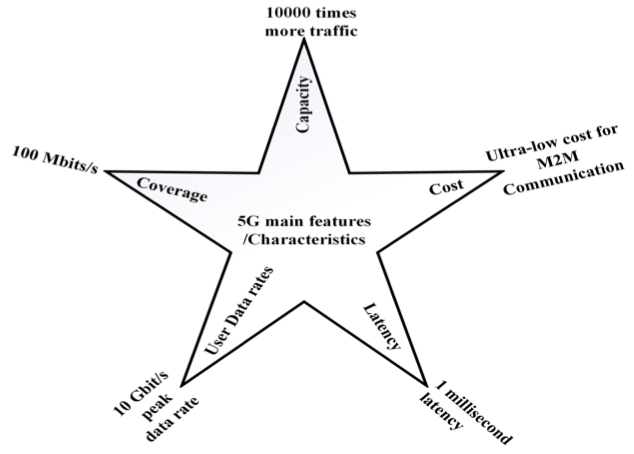}
  \centering
  \caption{The five key requirements of 5G \cite{ullah20195g}. }
  \label{fig:5G}
\end{figure}

In this article, we summarize some of the new research results of 5G in the field of autonomous vehicles. To qualify as a 5G network, it must follow some of the key requirements as summarized in Fig.\ref{fig:5G}. As shown in this figure, 5G's performance goals are high data rates, reduced latency, energy savings, reduced costs, increased system capacity, and large-scale device connectivity. These are also some of the characteristics amongst responsiveness, reliability, and others that the new era of autonomous cars seeks.

The faster data transmission will not only enable 5G networks to provide services for mobile phones but will also give a strong competition to the cable network providers, becoming the general home and office network providers. This will also facilitate the realization of IoT, smart home and smart cities along with accelerated development in autonomous vehicle applications. Currently, the companies that provide 5G wireless hardware and systems are Samsung, Ericsson, Qualcomm, Cisco, etc.

The main contribution of this survey is to provide state-of-the-art technological trends in terms of 5G communications, and especially 5G in autonomous vehicles. And also discuss the existing limitations and problems of existing cellular technologies. 

We further discuss our opinion on the design and applications of 5G in autonomous vehicles, our perspective on the future development of several elements of these studies, and the directions of research and implementation of vehicular communications which will gain traction in near future.

%% file: 2_Related_Work.tex
\section{Related Work}

The evolution of 5G has revolutionized a lot of technologies but an ideal 5G network for vehicular communication would consist of Safety Applications, Infotainment, IoT based applications, Security and Privacy. \cite{Katsaros2017} Current state-of-the-art networks promise dedicated short-range communications (DSRC) based on the IEEE 802.11p standard which also allows Hybrid Networks. But a lot of work is still required for congestion control, back-haul network, mobility management, and security to support vehicular networks. Research proposing the joint design of communication and control in self-driving automobiles have proven the use of 5G on miniature cars to successfully run autonomously on tracks designed with intersections. The cars run without colliding with each other with the help of a hybrid V2V and V2I communication. \cite{7848940}

Technologies like Blockchain and Artificial Intelligence are on the equal rise and when those are combined with 5G some promising results have been observed. \cite{8907054} Literature has research suggesting the use of content-centric networking (CCN) and blockchains that enable dynamic control of source reliability, legitimacy of the data transferred, security and integrity. This works better than the traditional transmission control protocol/Internet protocol (TCP/IP)\cite{8343866}. The self-driving vehicles collect information like local and global positioning, speed, performance, road and surrounding pictures, surface type, routes, accidents/colliding vehicles, etc. on an AI-based Base Station. the base station learns about the behavior of the road surfaces, vehicle interaction, pedestrian interactions, etc based on AI algorithms resulting in better and faster network architecture\cite{8663985}.

5G is an emerging technology that targets a lot of industry verticals has led to researchers employing the benefits of technology more and more in autonomous driving. The impact of 5G on autonomous driving can be broadly categorized into three parts:

The first is data transmission to the autonomous driving system on the vehicle. It is known that an autonomous driving system development is mainly divided into three parts: perception, decision-making, and control. The perception link requires a large amount of data collection. One of the important features of 5G is Proximity Service (ProSe) which provides information based on the locality of various devices, equipment, and services. Hence it is capable of providing high data transmission rates since it avoids going to the core of the network infrastructure to find the information. \cite{8255748}

The second is assisting with autopilot testing. At present, some 5G autonomous driving vehicle test roads have been opened in China, most of which are within 5 kilometers in length. By arranging 5G base stations and smart cameras on the test-beds, the 5G network coverage is finally completed. Strictly speaking, these test roads cannot directly affect autonomous driving vehicles, but they will indirectly promote automatic driving testing in the form of providing high-precision map data and enhancing information security\cite{doi:10.1080/13600869.2019.1696651}. 

 The third is the support provided by 5G for edge computing from the very beginning of the architecture design\cite{8123913}, and detailed designing of the network session management mechanism. They predict that in autonomous driving, edge computing can achieve ultra-low latency and improve business security\cite{7946184}. It is believed that when many vehicles are being controlled by cloud-based vehicle control systems\cite{7749210}, they can solve the challenge of limiting the induction of individual vehicles, which can lead to false warnings and deadlock situations.

These are also the foundation pillars for achieving the V2X (Vehicle-to-everything) communication standard that the majority of the industry aims to achieve. The V2X consists of V2V (Vehicle-to-vehicle), V2P (Vehicle-to-Pedestrian), V2I (Vehicle-to-Infrastructure) and V2N (Vehicle-to-Network) which requires a network to be able to communicate with these components to establish a 5G vehicular network. Establishing a network with vehicles will allow communication about the location and intended route amongst each other, resulting in safer driving. Connection with pedestrians is achieved via locally communicating with their mobile devices, which can result in instantaneous notification to the user as well as the vehicle in case of emergency or otherwise as well.

Similarly for other components and by establishing a successful V2X network, IoT can be realized at its full potential. V2X communication aims to deploy not just line-of-sight sensors like LIDAR, camera, radar, etc. but expects V2I, V2N, and V2V to perform accident or collide warnings, electronic charging and automatic park, payments at toll and speed limit notifications. This will result in better autonomous driving systems since it enables vehicles to communicate with garage, home appliances, nearby contacts, places, etc. and perform various tasks.

Generally, in the literature, surveys on 5G network including architecture \cite{agiwal2016next}, core network \cite{taleb2017multi}, caching \cite{ioannou2016survey} are available. In addition to state-of-the-art research surveys on the infrastructure of 5G systems, surveys on the latency reduction approach on Internet \cite{briscoe2014reducing}, cloud computing \cite{srivastava2016survey} are also presented. However, to the best of our knowledge, a more comprehensive review of the application of 5G systems to autonomous vehicles is lacking. So in the review, we focus on some related developments and cutting-edge research results on 5G in autonomous vehicles.

%% file: 3_latency.tex
\section{The applications of 5G in autonomous vehicles}

There are three service categories of 5G systems: enhanced mobile broadband(eMBB), massive machine-type communication (mMTC) and ultra-reliable and low latency communication (uRLLC) \cite{series2015imt}.

\begin{figure}[htbp]
  \includegraphics[scale=0.40]{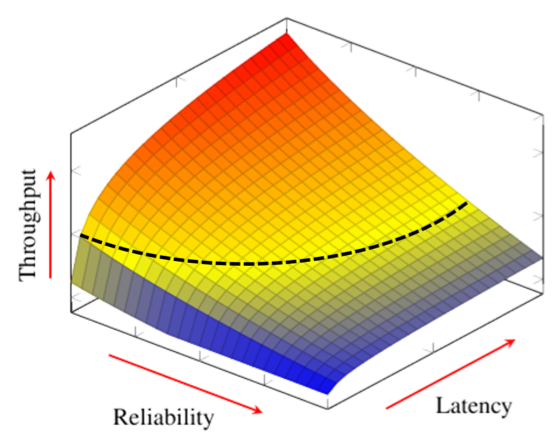}
  \caption{This figure shows the trade-off between throughout, latency and reliability [2].}
  \label{fig:tradeoff}
\end{figure}

The main objective of eMBB is to provide a higher data rate to users. eMBB mainly concerns the services that have high requirements for bandwidth, such as virtual reality(VR), augmented reality (AR) and high-resolution video streaming. The major emphasis in eMBB was placed on the improvement of system throughput, like average throughput, peak throughput, and cell-edge throughput, etc. Based on consensus, 5G not only provides personal mobile service (eMBB), but also mMTC, and latency/reliability critical services. In uRLLC, both the latency and reliability issues need to be addressed \cite{kumbhar2016survey}. 

For the mMTC, the main motivation is to increase the number of support devices and support low-cost devices. The major applications of mMTC service are logging, metering, and monitoring that need high connection density and ultra energy efficiency. uRLLC is a service that focuses on reducing end-to-end latency and increasing the robustness of data transmission, which is also one of the focus of this paper. uRLLC based applications are delay-sensitivity services, like autonomous driving, remote control and augmented reality, etc. 

Among the three service categories, uRLLC is the most critical and challenging factor in autonomous vehicle applications and hence we focus on it in the subsequent paper. It focuses on and addresses two key issues in autonomous vehicles: low latency and high reliability. 

As shown in Fig.\ref{fig:tradeoff}. \cite{soret2014fundamental}, there are trade-offs between latency, reliability, and throughput. The two requirements — low latency and high reliability — are conflicting. In the process of reducing the latency, usually, we use a short packet which will cause the channel coding gain to decrease severely. At the same time, in the process of improving the reliability, there are more costs of computing and communicating for redundancy, parity, and re-transmission, etc \cite{ji2017introduction}. Thus the latency is usually increased.

\subsection{Latency}

Latency is a critical factor in modern networked, real-time application systems, especially in a self-driving car or autonomous vehicle systems that emphasize safety and fast response. Intelligent and autonomous vehicles and road traffic control require low-latency and ultra-reliable communication. Autonomous vehicles need to coordinate with each other to do actions like platooning and overtaking etc \cite{campolo2017better}. For overtaking, the maximum E2E latency allowed for each message exchange is 10ms. For some video integrated applications in autonomous vehicles, the maximum latency is 50ms. Generally, for autonomous vehicles, it requires the latency of 10ms to 100 ms(max) with a packet loss rate of $10^{-3}$ to $10^{-5}$ \cite{parvez2018survey}.

The state-of-the-art technologies to resolve low latency problems can be divided into three solution aspects \cite{parvez2018survey}: Radio Access Network (RAN), core network, and caching. 

\begin{figure}[htbp]
  \includegraphics[width=\linewidth]{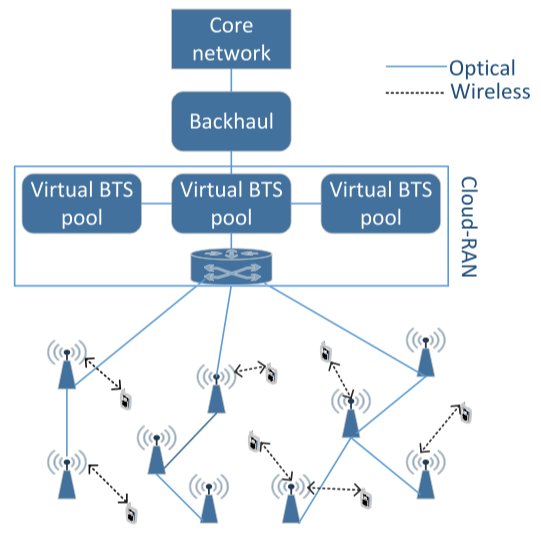}
  \caption{The architecture of CRAN in 5G networks. CRANs combine baseband processing units of a group of base stations into a central server retaining radio front end at the cell sides.}
  \label{fig:cran}
\end{figure}

Cloud Radio Access Network (CRAN), which is also an important development direction and an important design that can reduce latency, is introduced for 5G to reduce capital expenditure (CAPEX) and simplify the network management \cite{guizani2017cran}. The architecture of CRAN is shown in Fig.\ref{fig:cran}. CRAN requires connection links with the delay of less than 250 $\mu$s to support 5G low latency services.

In RAN, there are also some other refinements and approaches in different aspects been proposed, including control signaling, transmitter adapter, QoS/QoE differentiation, symbol detection, etc \cite{parvez2018survey}.

In addition to the refinements or enhancements in RAN, to meet the ultra-low latency vision in 5G, there are some other approaches proposed in the core network.

The new network entities software deﬁne network (SDN) and network virtualized function (NFV) are considered as the main design of the 5G core network \cite{agyapong2014design}. The SDN and NFV technologies in latency reduction are a focus in research of the 5G core network.

Mobility management in a core network based on SDN can potentially introduce some delays. Marquezan et al \cite{marquezan2016understanding} discussed the main contributors for processing delays in an SDN-based mobility management system. By implementing two proactive and reactive solutions for mobility management using Mininet and OpenFlow, it is observed that with high probability (almost 95\%) in the proactive mobility management system, the overall processing latency is around the median value. 

The NFV is proposed as a major entity for the 5G core network \cite{zhang2015architecture}. NFV removes the dependency on the hardware platform and makes ﬂexible deployment of EPC functions as well as the sharing of resources in RAN. This can reduce the E2E latency with improved throughput performance.

In caching, distributed and centralized caching with various trade-offs, cache placement and content delivery have been proposed.

These ideas are generally based on insufﬁcient capacity of backhaul links can also be considered as a bottleneck for low latency communication, in addition to the shortage of radio spectrum. 

As we talked previously, latency reduction is crucial for users’ QoS and QoE in 5G communications and autonomous vehicles' applications. Caching can be assumed as one of the promising candidate technologies to design a paradigm shift for latency reduction in 5G systems \cite{ioannou2016survey}. 

For the content caching aspect, researchers utilize diverse techniques to employ time intervals in which the network is not congested by filling appropriate data into the intervals. Shanmugam et al \cite{shanmugam2013femtocaching} proposed an approach to investigate the cache placement problem which achieved a considerable performance improvement for the users at reasonable QoS levels.

Generally, along with the huge capacity, massive connection density, and ultra-high reliability, 5G networks in autonomous vehicle systems and applications will need to support ultra-low latency. There are many different approaches been proposed in the domain of RAN, core network and caching.

\subsection{Reliability}
Before 5G emanated, the car-road collaboration project was mainly based on 4G technology. However, due to the relatively slow transmission speed of 4G-the highest is only 100 Mbit/s, resulting in a network rate delay of up to 50 milliseconds. At this speed, it is impossible to achieve real-time control for high-speed vehicles, so it is considered to restrict the development of autonomous driving. Get important factors.

In comparison, 5G networks can provide vehicles with millisecond-level ultra-low latency, transmission rates up to 10GB/s, as well as up to millions of connections per square kilometer and ultra-high reliability, helping vehicles perceive in remote environments. Breakthroughs have been made in key technologies such as information interaction and collaborative control, allowing vehicles to respond faster and drive more safely when faced with complex road conditions, which can be said to be an important cornerstone for achieving connected cars and autonomous driving.

Especially in some corner cases, there are always defects in relying only on the sensors on the bicycle. The autopilot fatal accident that occurred at Tesla last year is an important example. It addresses the limitations of car intelligence in terms of perception.

With the help of 5G technology, it can provide large bandwidth and low latency networks for cars and road infrastructure. It can provide high-end road perception and accurate navigation services for autonomous vehicles, enabling vehicles to detect longer distances and a wider range. The blind spots and dead corners in the driving process are well eliminated, the information transmission speed is faster, and the emergency response is faster, which greatly improves the safety and reliability of autonomous vehicles.

%% file: 5_Conclusion.tex
\section{Conclusion}
Based on the above discussions about 5G technology and autonomous vehicles, it is clear that with the closer integration of 5G and autonomous driving, 5G has a wider scope for autonomous driving. Amongst other key technologies like computer vision, robotics, analytic, etc. which are essential for the development and deployment of autonomous cars, 5G is equally important if not more. Without 5G, the advancement of other research areas would fail or not be realized completely because the systems would not be as reliable, latent, fast and other aspects as discussed in the above sections. 

If self-driving can be achieved entirely by intelligent bicycles, the role of 5G for autonomous driving may not be so important. However, the current problem is that the road to intelligent driving of bicycles by self-driving vehicles is difficult, and seeking external solutions is therefore on the agenda. When autonomous vehicles connect with the surrounding infrastructure, 5G can begin to work. 

Technology updates and iterations will drive considerable changes in the entire industry chain. The biggest change in the automotive industry is that cars become safer and transportation efficiency is further improved.  Therefore, the 5G network is of great significance for the popularization of driver-less technology, and even a necessary condition for the latter; at the same time, smart cars will also become the core applications of 5G networks. 5G technology will bring unprecedented communication changes to human society.